\documentclass{article}

\PassOptionsToPackage{numbers, compress}{natbib}
\usepackage[preprint]{neurips_2026}

\usepackage[utf8]{inputenc} % allow utf-8 input
\usepackage[T1]{fontenc}    % use 8-bit T1 fonts
\usepackage{hyperref}       % hyperlinks
\usepackage{url}            % simple URL typesetting
\usepackage{booktabs}       % professional-quality tables
\usepackage{amsmath}        % advanced math environments
\usepackage{amsfonts}       % blackboard math symbols
\usepackage{nicefrac}       % compact symbols for 1/2, etc.
\usepackage{microtype}      % microtypography
\usepackage{xcolor}         % colors
\usepackage{graphicx}
\usepackage{wrapfig}

\title{Beyond Routing: Characterising Expert Tuning and Representation in Vision Mixture-of-Experts}

\author{Gene Tangtartharakul \\
  School of Psychology\\
  University of Auckland\\
  Auckland, New Zealand\\
  \texttt{gene.tang@auckland.ac.nz}
  \And
  Katherine R. Storrs  \\
  School of Psychology\\
  University of Auckland\\
  Auckland, New Zealand\\
\texttt{katherine.storrs@auckland.ac.nz}}
\begin{document}

\maketitle

\begin{abstract}

Mixture-of-Experts (MoE) models are often interpreted by analysing which categories are routed to which experts. However, routing alone does not reveal what each expert actually encodes. We train sparsely-gated convolutional MoE models with a contrastive objective on natural images and characterise expert specialisation using tools from visual neuroscience. Extending from gating-level to expert-level analyses, we measure per-expert category separability, and per-expert tuning using the most exciting inputs. Extending from category-level to feature-level explanations, we interpret tuning via semantic dimensions derived from a dataset of human behavioural judgements (THINGS). Finally, we use tuning and representational similarity analysis to assess the stability of expertise-allocation across independent initialisations. We find that an animate–-inanimate distinction dominates expert partitioning, apparent from gating through to expert readout, and is stable across independently trained models. Although routing statistics suggest relatively sparse, categorical preferences, expert analyses reveal broader tuning to continuous visual and semantic dimensions that extend beyond category boundaries. Experts exhibit similar category-separability to one another, despite distinct feature tuning, demonstrating the explanatory benefits of moving beyond category-level analyses. Together, these results show that expert specialisation in vision MoEs extends well beyond category routing and is better understood by probing fine-grained expert-level tuning and representational structure.

\end{abstract}

\section{Introduction}

Mixture-of-Experts (MoE) has become an increasingly common architecture in deep neural networks across modalities and applications. Rather than relying on a single dense network, MoE models comprise multiple smaller, distinct subnetworks, or \textit{experts}, together with a gating network that learns to allocate inputs to appropriate subnetworks \citep{fedus_switch_2022, jacobs_adaptive_1991, shazeer_outrageously_2017, jordan_hierarchical_1994, mu_comprehensive_2026}. After training, experts are often found to capture different components of a task. Modern implementations of MoE architectures typically induce sparsity \citep{fedus_switch_2022, han_vimoe_2024, riquelme_scaling_2021, shazeer_outrageously_2017, shen_scaling_2023}, such that only $k$ experts (out of $E$) are activated for a given input, where $k \ll E$. MoE design naturally encourages expert specialisation by partitioning a task into disentangled subspaces that can be associated with different experts, thereby aiding interpretability \citep{gan_mixture_2025, yang_mixture_2025}. Understanding what a model has learned and why is crucial for revealing internal representations, diagnosing failures, and identifying targets for alignment and model improvement. MoE architectures provide an explicit inductive bias toward interpretability, as each expert can be inspected in isolation, making it possible to identify which input features or subtasks activate which expert \citep{pavlitska_sparsely-gated_2023, mustafa_multimodal_2022, wang_moiie_2026}. Lightweight gating also improves accessibility into the model's internal routing and decision-making processes.

Several studies have examined MoE interpretability \citep{mustafa_multimodal_2022, dutt_exploiting_2025, han_vimoe_2024, riquelme_scaling_2021, wang_moiie_2026, videau_mixture_2025, cai_long-tailed_2026, chaudhari_moe_2026, lo_closer_2025, olson_probing_2025, ying_beyond_2025, nikolic_exploring_2025}. In language MoEs, expert specialisation often reflects syntactic structure, semantic structure, or a combination of both \citep{antoine_part--speech_2024, zoph_st-moe_2022, shazeer_outrageously_2017, olson_probing_2025, xue_openmoe_2024}, with studies commonly reporting specialisation for parts of speech \citep{antoine_part--speech_2024, zoph_st-moe_2022, shazeer_outrageously_2017}. In vision MoEs, experts are often reported to specialise according to semantic categories (e.g., animals and plants) \citep{pavlitska_sparsely-gated_2023, riquelme_scaling_2021, nikolic_exploring_2025}, object size \citep{pavlitska_sparsely-gated_2023}, or spatial region \citep{han_vimoe_2024, pavlitska_design_2026}, depending on the training objective \citep{pavlitska_design_2026, videau_mixture_2025}. In multimodal MoEs, experts may likewise partition inputs within or across modalities \citep{shen_scaling_2023}.

Across MoE studies, expert specialisation is almost exclusively probed through gating information. In language MoEs, routing decisions are examined through token-to-expert assignments and usage frequency \citep{antoine_part--speech_2024, zoph_st-moe_2022, shazeer_outrageously_2017}; in vision MoEs, through gating weights and expert-by-class heatmaps \citep{pavlitska_sparsely-gated_2023, riquelme_scaling_2021, nikolic_exploring_2025}; and in multimodal MoEs, through modality-usage charts \citep{shen_scaling_2023}. As a result, existing work largely treats routing decisions as a proxy for specialisation rather than directly examining the experts themselves, leaving the representations and feature-tuning profiles experts learn poorly understood.

Analogous questions have been studied extensively in visual neuroscience. The occipitotemporal cortex (OTC) contains functionally specialised modules with well-documented selectivity for colour \citep{mckeefry_position_1997}, shape \citep{malach_object-related_1995, grill-spector_cue-invariant_1998}, faces \citep{kanwisher_fusiform_2006, kanwisher_fusiform_1997}, places \citep{epstein_cortical_1998}, body parts \citep{downing_cortical_2001}, and words \citep{hannagan_origins_2015, lochy_selective_2018}. Although these regions are traditionally described as category-selective, accumulating evidence favours a more distributed account: rather than being tuned to a single category, representations within these regions are broadly sensitive to finer-grained, behaviourally relevant dimensions \citep{contier_distributed_2024, dyck_multidimensional_2025, lugtmeijer_visual_2025, ritchie_rethinking_2026}. This motivates a broader interpretation of expert specialisation in MoEs, and suggests that the neuroscientific toolkit for characterising distributed representations may offer a useful lens for probing vision MoEs. We return to this point in the next section.

In this work, we build a sparsely gated contrastive Mixture-of-Experts convolutional neural network and characterise expert specialisation using tools drawn from visual neuroscience. Our contributions are:
\begin{enumerate}
    \item \textbf{Beyond routing.} We extend MoE interpretability from gating-level statistics to expert-level tuning and population-level representational geometry, showing that routing alone systematically mischaracterises what experts encode.
    \item \textbf{Beyond categories with continuous perceptual dimensions.} We use the THINGS dataset \citep{hebart_things_2019} and its 66 human behaviour-derived visual and semantic dimensions \citep{hebart_revealing_2020} as an out-of-distribution probe, disentangling expert tuning from the categorical structure of the training set and recovering interpretable, continuous tuning profiles.
    \item \textbf{Representational stability.} Across independently trained models ($E \in \{4, 8, 16\}$, 10 seeds each), we identify representational motifs that re-emerge reliably using second-order RSA and silhouette-based cluster selection.
\end{enumerate}
Together, these results reframe expert specialisation in vision MoEs as sparse tuning to continuous, behaviourally relevant dimensions rather than to discrete categorical classes.

%% Related Work Start %%

\section{Related Work}

\paragraph{Interpretability of Vision and Multimodal MoEs.} MoE layers have been integrated with various vision backbones, including ViT \citep{riquelme_scaling_2021}, VAE \citep{nikolic_exploring_2025}, and CNN \citep{pavlitska_sparsely-gated_2023}. Across these works, expert specialisation is predominantly characterised through routing dynamics as a proxy: per-class averages of routing weights or logits \citep{wu_residual_2022}, routing frequency \citep{han_vimoe_2024, vashkelis_hi-moe_2026}, and visualisations of gating patterns via low-dimensional embeddings or per-expert exemplars \citep{pavlitska_sparsely-gated_2023, nikolic_exploring_2025}. In multimodal MoEs, one study extended beyond routing to within-expert probing \citep{ying_beyond_2025} by adapting the Model Utilisation Index (MUI) to MoE architectures. This measure quantifies the proportion of neurons within each expert activated for task completion. Nevertheless, such a probing technique remains a functional measure of recruitment, as it identifies which neurons are engaged but not what they encode, leaving expert tuning and representational structure unexplored.

\paragraph{Functional Analyses of Cortical Visual Areas.} In neuroscience, various tools have been developed to characterise the tuning properties of cortical areas across multiple scales. Tuning at the level of individual neurons or neural populations can be probed using techniques such as Most Exciting Inputs (MEIs), which have long been used in animal \citep{adrian_discharge_1928, hartline_response_1938} and human neurophysiology, and more recently in deep neural networks \citep{walker_inception_2019}, to identify the stimuli that maximally drive individual neurons or neural populations. MEIs can be selected from natural image sets or generated synthetically to identify optimal inputs. At a broader scale, the structure of stimulus representations has been characterised through representational geometry --- how stimuli cluster in population response space \citep{kriegeskorte_neural_2021}. Multivariate analyses, for instance decoding or category separability, can probe which feature properties, such as category structure, are entangled or disentangled within the representational manifold. Representational Similarity Analysis (RSA) can likewise be used to quantify similarities and differences in the internal organisation of stimulus representations across systems \citep{kriegeskorte_representational_2013}. 

\paragraph{Properties of Expert Areas in the Brain.} Traditionally, expert areas have been understood as regions that show selectivity for specific categories, such as the fusiform face area (FFA) \citep{kanwisher_fusiform_1997, kanwisher_fusiform_2006}, the parahippocampal place area (PPA) \citep{epstein_cortical_1998}, the extrastriate body area (EBA) \citep{downing_cortical_2001}, and the visual word form area (VWFA) \citep{petersen_positron_1988, dehaene_unique_2011}. Despite being most strongly activated by their preferred categories, studies have shown that these regions are broadly tuned to a range of features, spanning from low-level properties such as image statistics and eccentricity \citep{rice_low-level_2014, hasson_eccentricity_2002, arcaro_retinotopic_2009} to mid-level dimensions such as animacy and size \citep{long_mid-level_2018, thorat_nature_2019, konkle_tripartite_2013}. More recent evidence further suggests that mid-level visual and semantic dimensions, such as `spikiness', `metallic-ness', and `plant-relatedness', explain variance in neural responses in visual cortical regions better than categorical labels alone \citep{contier_distributed_2024}, highlighting the role of behavioural relevance in human cortical representations. This suggests that expert selectivity cannot be adequately understood as purely category-selective, nor as consisting solely of sparse peaks in representational space. On this view, expertise is better understood as a response peak embedded within a broader distributed and continuous representation.

%% Related Work Ends %%

%% Method Starts %%

\section{Sparse Contrastive MoE-CNN}

\begin{wrapfigure}{r}{0.6\linewidth}
  \centering
  \includegraphics[width=0.6\linewidth]{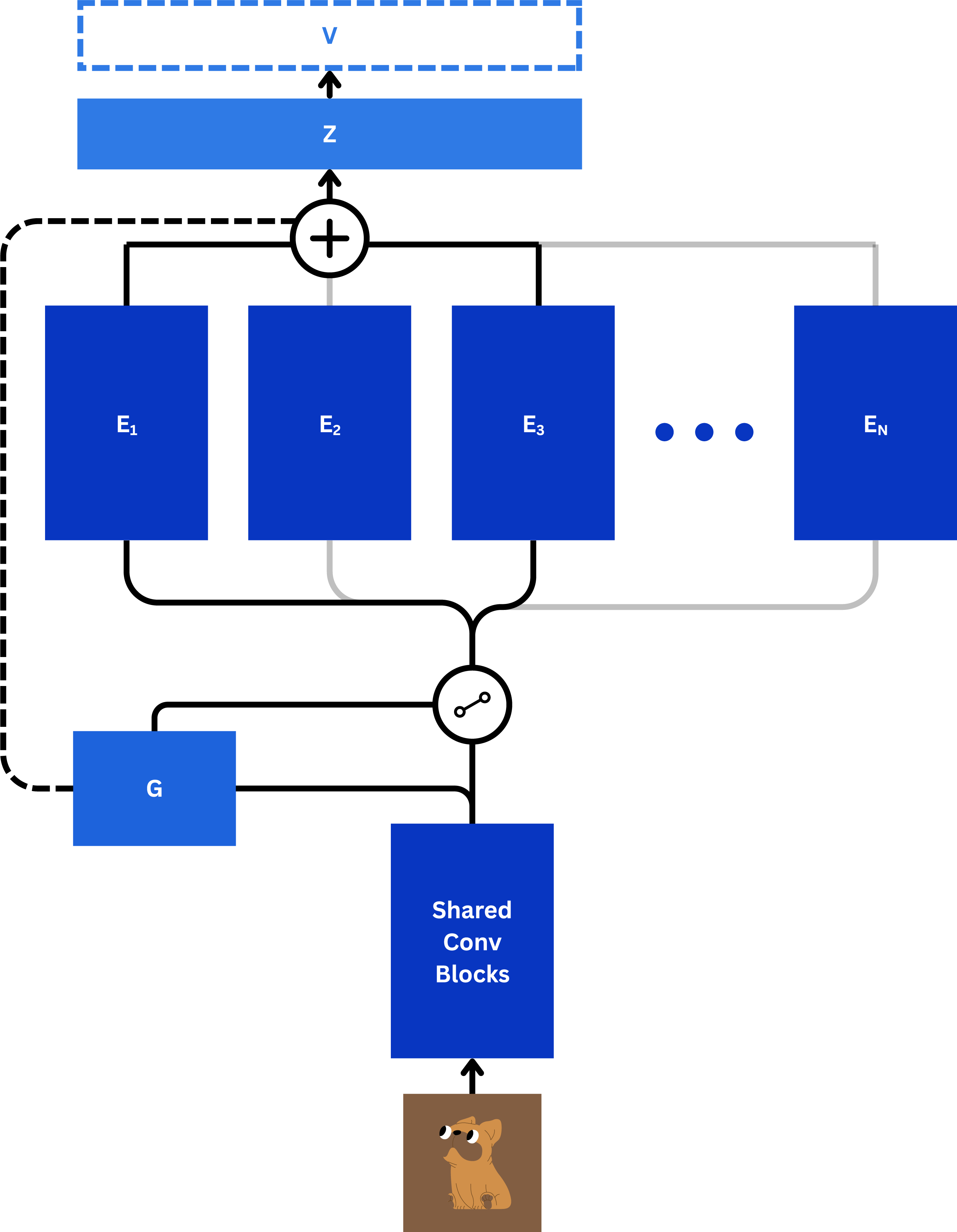}
  \caption{Overview of the model architecture. The input image is first processed by a shared backbone, after which the gating network uses the latent representation to select the top-$k$ experts that further process the sample. The final embedding is the weighted sum of the activated experts' readouts. During training only, a projection head is appended to this embedding for contrastive learning (dashed box).}
  \label{fig:mainfig_architecture}
  \vspace{-0.8\baselineskip}
\end{wrapfigure}

\paragraph{Model Architecture.} Our design takes inspiration from the biological brain, with hierarchical processing constrained by progressively expanding receptive fields and spatial locality. We use a standard ResNet \citep{he_deep_2015} backbone of five convolutional blocks followed by a three-layer MLP readout. We adapt this into an MoE by inserting a gating network after the first three convolutional blocks (the \textit{shared blocks}). The position of the gating layer is known to influence both routing decisions and the features experts learn \citep{videau_mixture_2025, pavlitska_sparsely-gated_2023}. We chose this position to allow moderate shared processing before specialisation emerges (Figure~\ref{fig:mainfig_architecture}).

After the three shared blocks, the remaining two convolutional blocks are replicated into $E$ independent experts, of which only the top-$k$ continue processing each input, as determined by the gating network. Each expert retains its own readout block. To ensure that any observed specialisation does not arise merely from increasing the number of experts, we control for the total number of model parameters: as $E$ increases, each expert's channel width is divided by $\sqrt{E}$, keeping the total parameter count approximately constant across models ($\sim 46$M; range: 45.8--46.1M for $E \in \{4, 8, 16\}$). A consequence of this design is that, as $E$ increases, each expert has progressively less individual capacity, encouraging experts to develop complementary rather than overlapping tuning to similarly cover the feature space. This loosely echoes the resource constraints faced by biological neural populations, where broad feature coverage must be achieved through complementary tuning across units rather than within them.

For each input, the model thus outputs an embedding vector $\mathbf{z} \in \mathbb{R}^{\lfloor 256/\sqrt{E} \rfloor}$. This embedding is computed as the gating-weighted sum of the selected experts' readouts:
\begin{equation}
  \mathbf{z} = \sum_{i \in \mathcal{K}} g_i \, \mathbf{z}_i
\end{equation}
where $\mathcal{K}$ is the set of top-$k$ selected experts, $g_i$ is the gating logit, and $\mathbf{z}_i$ is the readout of expert $i$. During training, a projection head $\operatorname{Proj}(\cdot)$, implemented as a two-layer MLP with batch normalisation and ReLU, maps $\mathbf{z}$ to a projected embedding $\mathbf{v} = \operatorname{Proj}(\mathbf{z}) \in \mathbb{R}^{128}$. This design allows the projection head to be discarded after training while preserving the quality of $\mathbf{z}$ for downstream analyses \citep{xue_investigating_2024}.

\paragraph{Gating Network.} The gating network operates on the latent $\mathbf{h} \in \mathbb{R}^{C \times H \times W}$ produced by the last shared convolutional block, and outputs expert-selection probabilities $G(\mathbf{h}) = [g_1(\mathbf{h}), \dots, g_E(\mathbf{h})]$. It comprises two parallel branches sharing the architecture
\begin{equation}
    \phi(\mathbf{h}; \theta) = \operatorname{FC}\!\left(\operatorname{GAP}\!\left(\operatorname{ReLU}\!\left(\operatorname{BN}(W_{\text{conv}} * \mathbf{h})\right)\right)\right) \in \mathbb{R}^{E},
\end{equation}
but with independent parameters $\theta_g, \theta_n$. Inspired by~\citep{pavlitska_sparsely-gated_2023}, each branch combines a $3\!\times\!3$ convolution, global average pooling, and a fully connected layer, allowing detailed feature information to influence gating. We additionally use noisy gating~\citep{shazeer_outrageously_2017} during training to encourage load balancing. Formally,
\begin{equation}
\begin{aligned}
    \boldsymbol{\ell}_g &= \phi(\mathbf{h}; \theta_g), \quad \boldsymbol{\ell}_n = \phi(\mathbf{h}; \theta_n), \\
    H(\mathbf{h})_i &= \ell_{g,i} + \mathcal{N}(0, 1) \cdot \operatorname{Softplus}(\ell_{n,i}), \\
    G(\mathbf{h}) &= \operatorname{Softmax}\!\left(\operatorname{KeepTopK}(H(\mathbf{h}), k)\right), \\
    \operatorname{KeepTopK}(\mathbf{v}, k)_i &=
    \begin{cases}
        v_i & \text{if } v_i \text{ is among the top-}k \text{ entries of } \mathbf{v}, \\
        -\infty & \text{otherwise.}
    \end{cases}
\end{aligned}
\end{equation}
Here, $\boldsymbol{\ell}_g \in \mathbb{R}^E$ is the deterministic logit vector and $\boldsymbol{\ell}_n \in \mathbb{R}^E$ produces input-dependent noise scales (via Softplus) added to $\boldsymbol{\ell}_g$ before top-$k$ sparsification.

\paragraph{Losses.} We used the Normalised Temperature-Scaled Cross-Entropy (NT-Xent) loss. We chose a self-supervised contrastive learning objective because it does not constrain representations to predefined object categories. Each image instance is treated as distinct, with only two augmented views of the same instance considered similar \citep{prince_contrastive_2024}. This allows data-driven visual-feature specialisation to emerge rather than baking categorical structure into the training objective. Here, the loss is calculated in the projected space $\mathbf{v}$, encouraging augmented images derived from the same source image to have higher cosine similarity in the final embedding space, while images from different sources are encouraged to be dissimilar. We used a temperature of $\mathcal{\tau} = 0.5$. To discourage expert collapse, we penalise the squared coefficient of variation ($\mathrm{CV}^2$) of per-expert importance $\mathcal{I}$~\citep{shazeer_outrageously_2017}, where $\mathcal{I}_i = \sum_{\mathbf{x}} g_i(\mathbf{h}(\mathbf{x}))$ is the total gating weight assigned to expert $i$:
\begin{equation}
    \mathcal{L}_{\text{importance}} = w_{\text{importance}} \cdot \left(\frac{\mathrm{Std}(\mathcal{I})}{\mathrm{Mean}(\mathcal{I})}\right)^2,
\end{equation}
with $w_{\text{importance}} = 0.1$.

We observed that, together with noisy top-$k$ gating, this loss produces reasonable and consistent load balancing across experts. Additionally, although there is no inductive bias forcing augmented views of the same image to pass through the same set of $k$ experts, this pattern emerged naturally (Appendix B.1).

\paragraph{Training.} We trained our sparsely gated contrastive MoE-CNN on the STL10 unlabelled dataset ($96 \times 96$; 100k images) \citep{coates_analysis_2011} and validated on its labeled split (5k images) for 100 epochs. Our contrastive training follows SimCLR: given an origin image $\mathbf{x}$, two augmented views $\mathbf{x}_i$ and $\mathbf{x}_j$ are generated, and the model is trained to maximise similarity between positive pairs (same origin) and minimise similarity between negative pairs (different origins) in the projected space $\mathbb{R}^{d'}$. We trained three model configurations with varying numbers of experts, $E \in \{4, 8, 16\}$, with top-$k = 2$. See Appendix A.1 for further details. We held out the STL10 test split as a part of testing later in the experiments. 

\paragraph{Inference Dataset: THINGS.} The THINGS dataset contains 26,107 high-quality, manually curated images spanning 1,854 diverse object concepts \citep{hebart_things_2019}. Unlike STL10, whose narrow category set and prototypical image structure conflate categorical and low-level visual tuning, THINGS provides a systematically and densely sampled set of naturalistic object images that span a broad and diverse semantic space. We therefore used THINGS as an out-of-distribution (OOD) inference dataset to better disentangle the visual properties to which each expert responds. Crucially, THINGS is accompanied by the THINGS Similarity dataset \citep{hebart_revealing_2020}, which provides a representative image for each of the 1,854 concepts along with 66 human behaviour-derived dimension embeddings. The embeddings were derived from a sparse non-negative dimensionality reduction applied to $\sim$4.7M triplet odd-one-out human similarity judgments collected on 1,854 natural object concepts from the THINGS database. The resulting dimensions were labelled with human descriptors. We exploit these embeddings to quantitatively interpret each expert's tuning preferences beyond categorical labels.

%% Method Ends %%

\section{Results}

Throughout this section, we focus our detailed analyses on a subset of model configurations, for instance $E = 4$ or $E \in \{4,8\}$ to keep expert characterisation tractable. We report $E = 16$ where it adds value (e.g., Figure~\ref{fig:supplementary_mei}) and in the appendix otherwise.

\begin{figure}[t]
  \centering
  \includegraphics[width=\linewidth]{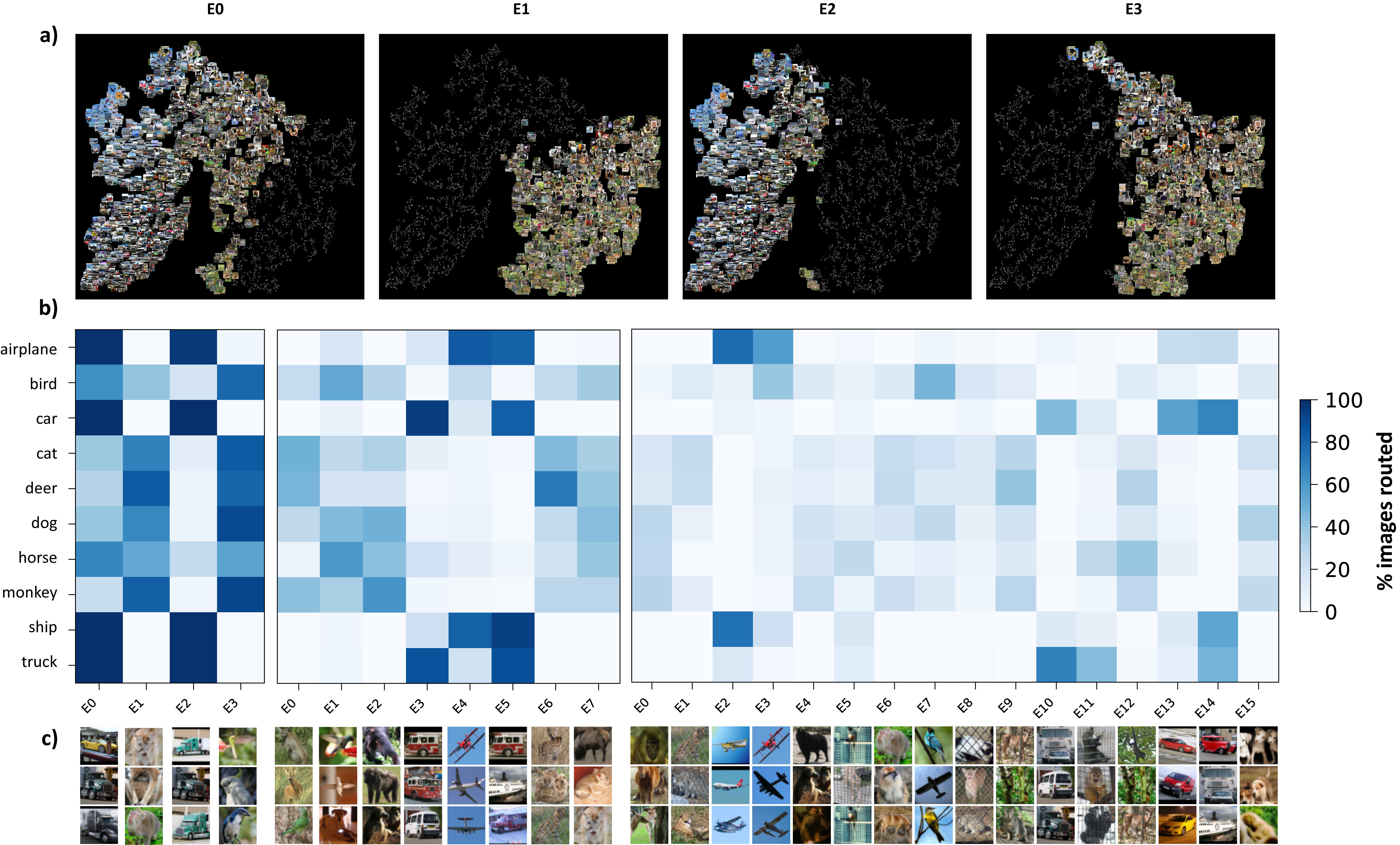}
  \caption{Expert specialisation through the lens of gating on the held-out STL10 test set (8k images; 800 images per class). (a) t-SNE plots showing the gating partition of an example four-expert model. Each column displays the subset of images routed to one expert. (b) Heatmap showing the proportion of images from each class routed to each expert across three models of 4, 8, and 16 experts. (c) For each expert shown in (b), the top three images that maximally activate its gating logits.}
  \label{fig:mainfig_routing_freq}
\end{figure}

\paragraph{Probing Specialisation Through Routing.} Standard methods examine expert specialisation through routing frequency and gating weights \citep{pavlitska_sparsely-gated_2023, antoine_part--speech_2024, shen_scaling_2023, zoph_st-moe_2022, shazeer_outrageously_2017}. Using the held-out STL10 test set, we analysed the proportion of each of the 10 image classes routed to each expert (Figure~\ref{fig:mainfig_routing_freq}). Routing profiles differed across experts, with the $E = 4$ model showing pairs of experts with similar routing proportions. While such pairing was less pronounced in models with more experts, a recurring pattern emerged: individual experts tended to receive inputs that were predominantly either animate or inanimate. We further visualised the top three images corresponding to the highest gating logits (pre-softmax) for each expert, which made this distinction salient. We also observed that the highest-logit images for a given expert commonly belonged to the same object category. On the basis of routing analyses alone, one might conclude that experts specialise by coarse semantic domain, for example trucks, birds, or planes.

\begin{figure}[t]
  \centering
  \includegraphics[width=\linewidth]{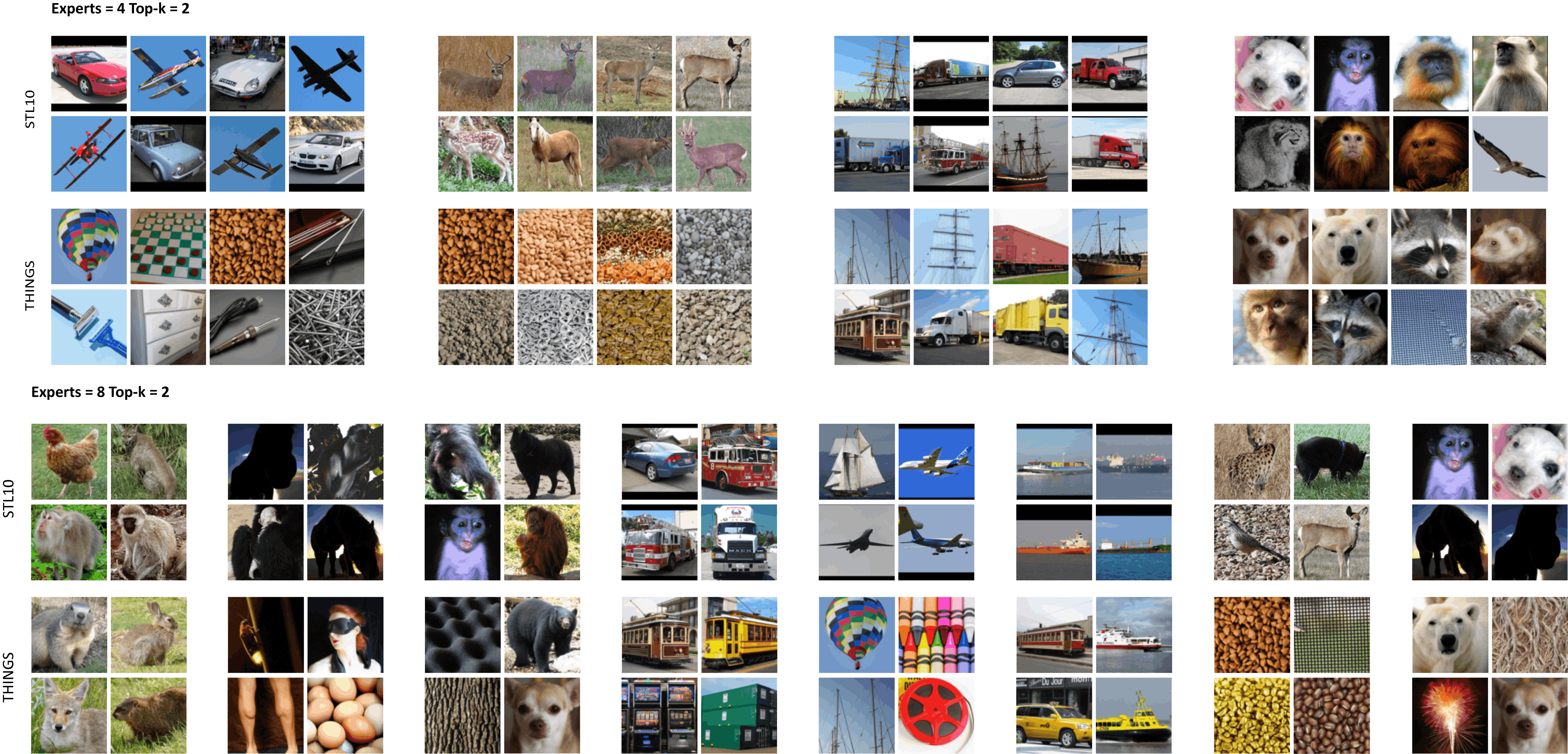}
  \caption{Images eliciting the highest response magnitudes at the readout layer for each expert. The upper half displays the top eight images for the four-expert model, and the lower half displays the corresponding images for the eight-expert model (See $E = 16$ further in the appendix Figure~\ref{fig:supplementary_mei}). Within each half, the upper panel shows maximal-response images from the held-out STL10 test set, whereas the lower panel shows maximal-response images from the THINGS dataset.}
  \label{fig:mainfig_meis}
\end{figure}

\paragraph{Expert-Level Tuning.} Beyond gating, we examined each expert's tuning using most exciting inputs (MEIs), which reveal the features to which an expert is most responsive. We did this by visualising the top images that elicited the largest readout response magnitude for each expert, quantified as the $L_2$ norm of the readout embeddings, $\|\mathbf{z}_i\|_2$. To compute this for every image regardless of gating, we bypassed the router and forced all images through each expert. We first identified MEIs from the held-out STL10 test set. Top images from STL10 suggested that experts preferentially responded to certain semantic classes, but given STL10's limitations, we repeated the analysis using THINGS to better disentangle each expert's true tuning. Among experts that appeared to share the same categorical preference under STL10, THINGS revealed that some were in fact responding to lower-level image properties rather than the category per se. For instance, one expert that appeared deer-selective under STL10 instead preferred grainy, high-spatial-frequency image statistics shared across those images when probed with THINGS (Figure~~\ref{fig:mainfig_meis}).

To quantify this more rigorously, we leveraged the THINGS Similarity dataset, which provides 66 visual and semantic dimensions, derived from human judgements, for each of 1,854 object concepts, each exemplified by one image. We passed all 1,854 concept images through the model and, for each expert, recorded both its gating logit (a scalar from the router) and its readout magnitude $\|\mathbf{z}\|_2$ (the $L_2$ norm of the $\lfloor 256/\sqrt{E} \rfloor$-dimensional readout embedding computed by forcing the image through that expert, as done above). We then fit nested cross-validated non-negative Lasso regressions to predict each of these targets from the THINGS embeddings (see Appendix A.2). This allowed us to recover a sparse and interpretable set of top tuning dimensions describing what the gating network routes to each expert, and what each expert responds most strongly to. In Table~\ref{tab:things_loadings}, we report the three dimensions with the largest regression coefficients as labels for each expert's tuning profile in a four-expert model (see the expanded tables for the eight- and sixteen-expert models in Appendix Table~\ref{tab:supplementary_things_loadings}). 

\begin{figure}[t]
  \centering
  \includegraphics[width=\linewidth]{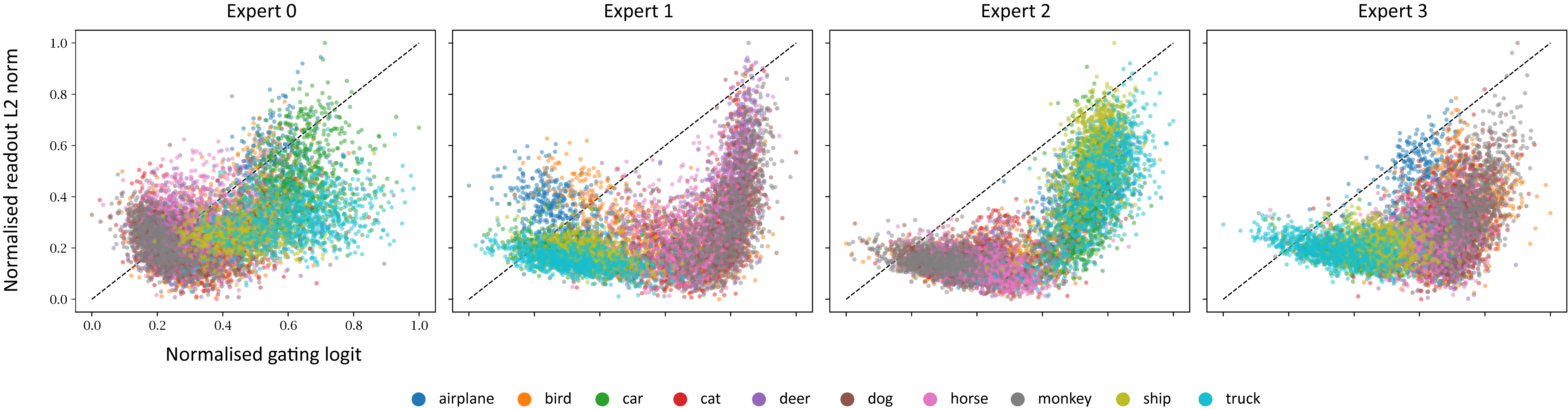}
    \caption{Scatterplot showing the relationship between gating logits and readout activation norms across the held-out STL10 test set. Min--max-normalized gating logits are plotted against normalized readout response magnitudes for each expert in a four-expert model. Points are coloured by image class. The dashed diagonal indicates the relationship expected if increases in gating logits always corresponded to increases in readout magnitude.}
    \label{fig:mainfig_gating_vs_readout}
\end{figure}

Strikingly, experts that initially appeared category-tuned (Figure~\ref{fig:mainfig_routing_freq}) were instead preferentially tuned to shape, colour, or texture features that do not necessarily align with categorical boundaries. We also found that gating logits and readout response magnitudes were not linearly related for any of the experts (Figure~\ref{fig:mainfig_gating_vs_readout}; See Figure~\ref{fig:supplementary_gating_vs_readout} for the eight-expert and sixteen-expert models), indicating that what routes to an expert is not what the expert most strongly encodes. 

Together, these analyses show that gating-level and readout-level investigations can lead to different conclusions about expert specialisation. This divergence likely arises because each expert further transforms its gating-partitioned inputs, discovering and amplifying features that need not align with those used for routing.

\begin{table}[t]
  \caption{Loadings of behaviour-derived dimensions on each expert gating and response-magnitude profile for the four-expert model. Each row reports the expert index, the $r^2$ from a five-fold cross-validated non-negative Lasso regression, and the top three dimensions with their corresponding coefficients.}
  \label{tab:things_loadings}
  \centering
  \scriptsize
  \setlength{\tabcolsep}{3pt}
  \resizebox{\linewidth}{!}{%
  \begin{tabular}{lcclclclc}
    \toprule
    Expert & $r^2$ Mean & $r^2$ Std  & Feature 1 & Weight & Feature 2 & Weight & Feature 3 & Weight \\
    \midrule
    \multicolumn{9}{l}{\textit{$E = 4$, k = 2 — Gating}} \\
    \midrule
    0 & .475 & .015 & house-related/furnishing-related & 0.276 & electronics/technology & 0.223 & metallic/artificial & 0.211 \\
    1 & .531 & .013 & animal-related & 0.496 & powdery/earth-related & 0.345 & food-related & 0.342 \\
    2 & .538 & .017 & house-related/furnishing-related & 0.281 & transportation-/movement-related & 0.218 & electronics/technology & 0.172 \\
    3 & .416 & .018 & animal-related & 0.284 & food-related & 0.247 & body part-related & 0.160 \\
    \midrule
    \multicolumn{9}{l}{\textit{$E = 4$, k = 2 — Readout}} \\
    \midrule
    0 & .144 & .033 & electronics/technology & 0.138 & black & 0.118 & transportation-/movement-related & 0.116 \\
    1 & .125 & .058 & fine-grained pattern & 0.341 & repetitive/spiky & 0.223 & animal-related & 0.191 \\
    2 & .200 & .036 & transportation-/movement-related & 0.222 & long/thin & 0.148 & electronics/technology & 0.133 \\
    3 & .144 & .029 & animal-related & 0.182 & body part-related & 0.150 & flying-/sky-related & 0.090 \\
    \bottomrule
  \end{tabular}%
  }
\end{table}

\begin{figure}[!ht]
  \centering
  \includegraphics[width=\linewidth]{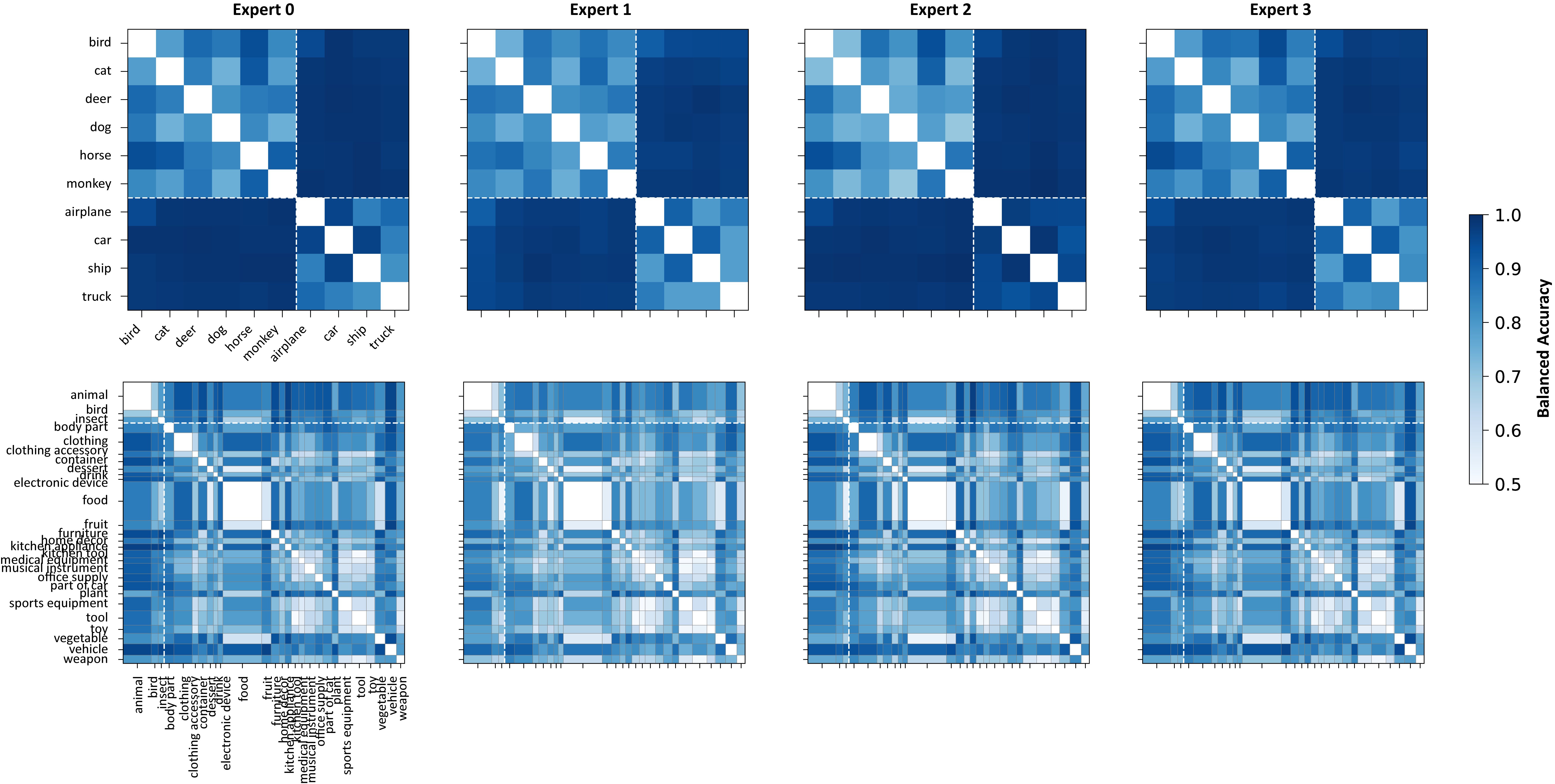}
  \caption{Category-separability heatmaps for an example four-expert model. The heatmaps depict pairwise category separability at readout. Five-fold cross-validated logistic regression was used to estimate classification accuracy (chance level $= 0.5$) for each category pair in the (top) held-out STL10 test set and the (bottom) full THINGS dataset. The dashed line indicates the animacy--inanimacy boundary. In the bottom panel, cell sizes scale with the number of image concepts assigned to each category. Balanced accuracy is reported.}
  \label{fig:mainfig_separability}
\end{figure}

\paragraph{Separability Analysis.} Another lens through which expert representations can be understood is representational geometry, which captures the pairwise distance structure between images as encoded by the population response. Unlike MEIs, which provide single-image responses, multivariate analyses characterise the information extractable at the population level, for instance, the linear separability of image categories. Having observed animacy as a primary distinction through THINGS dimensional loadings, we asked whether expert representations go beyond animate--inanimate distinctions toward finer-grained category individuation (i.e. distinguishing categories within the same animacy class). We passed all samples from the held-out STL10 test set through a four-expert model and recorded activations at readout. Pairwise logistic regression was trained on these activations to assess between-category discriminability (see Appendix A.3). 

As expected, all experts discriminated animate from inanimate categories. Experts also distinguished categories within the same animacy class, albeit to a lesser extent. This pattern held both within the trained distribution (the held-out STL10 test set) and on the OOD THINGS dataset (Figure~\ref{fig:mainfig_separability}). Notably, representational geometries were strikingly similar across experts despite their divergent MEIs, dissociating representational geometry from neuronal implementation \citep{lin_neuronal_2023}. That is, functionally equivalent category-level distance structures can be realised through distinct population-level response patterns. This underscores the need to probe expert representations at multiple scales.

\paragraph{Representational Stability Across Instantiations.}
Lastly, we quantified whether expert specialisations re-emerged consistently across independent model initialisations. For each configuration $E \in \{4, 8, 16\}$ with $k = 2$, we trained 10 models from different random seeds. To identify representations that persisted across runs, we applied representational similarity analysis (RSA) to all experts jointly. For each expert, we extracted readout-layer activations to 1,854 reference images from the THINGS dataset and computed a first-order representational dissimilarity matrix (RDM), $\mathrm{RDM} \in \mathbb{R}^{{1854} \times 1854}$. Each RDM entry captures the pairwise Euclidean distance between two image representations (See rationale in Appendix A.4). We then took the upper-triangular vector of each RDM and constructed a second-order similarity matrix across all (model $\times$ expert) pairs using Spearman correlation, $r_s$. This is a standard second-order RSA measure that captures how similar experts' representational geometries are to one another, independent of which model they came from. 

We applied agglomerative hierarchical clustering with average linkage to this second-order similarity matrix to identify groups of experts with consistent representational structure across runs. Rather than assuming that the number of clusters should equal the number of experts, we performed a grid search across 2 to 15 clusters and selected the number of clusters that maximised the silhouette score (range $[-1, 1]$) (See Figure~\ref{fig:supplementary_scree} for silhouette score plot), which measures whether each expert is more similar to its own cluster than to the nearest neighbouring cluster. Under this interpretation, large clusters reflect representational motifs recovered reliably across initialisations, whereas small clusters indicate rarer or less stable specialisations. Finally, we visualised the shared features within each cluster using the most exciting inputs (MEIs) drawn from example models.

\begin{figure}[t]
  \centering
  \includegraphics[width=\linewidth]{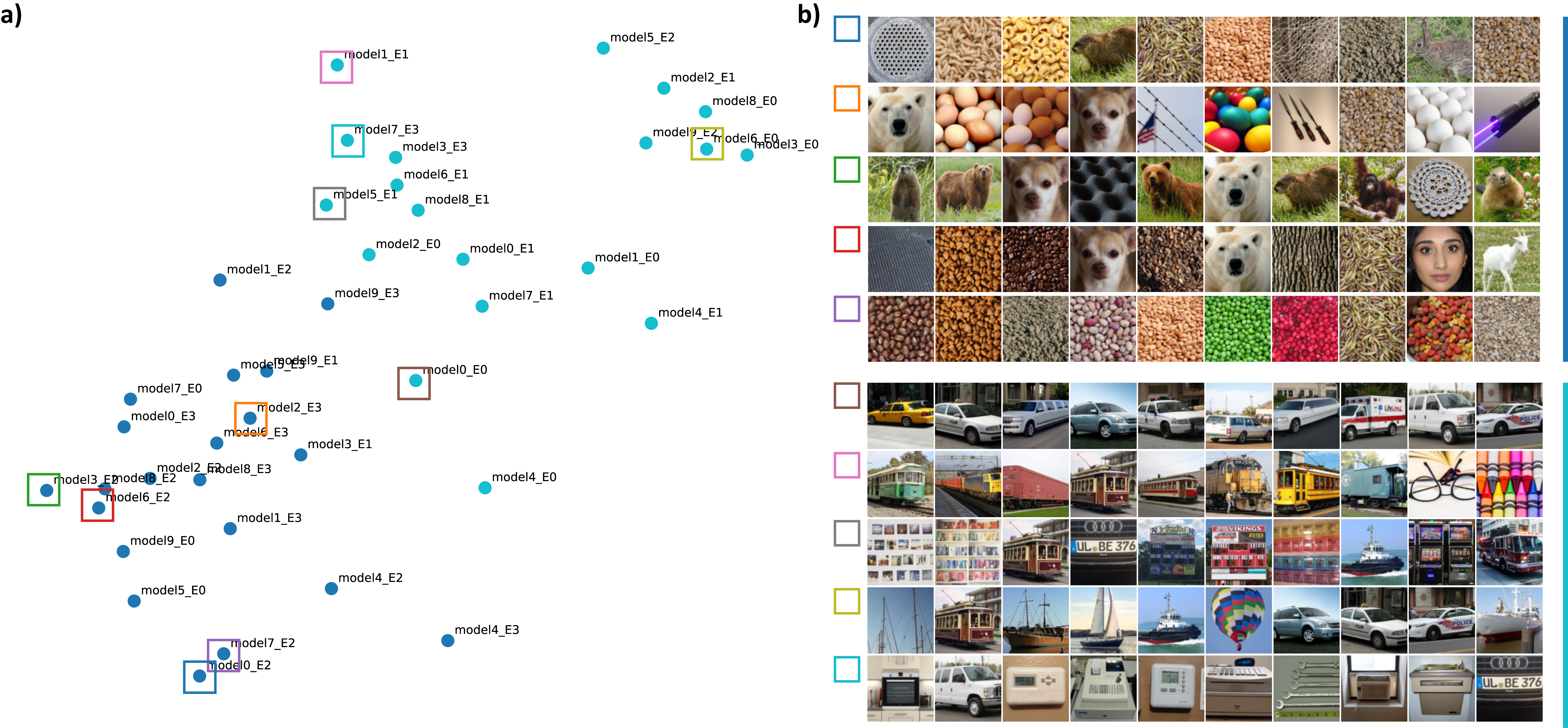}
  \caption{Consistency of expert specialisations across 10 independently trained model instantiations for four-expert model. Two clusters best capture the grouping structure of learned representations. a) Two-dimensional projection of each (model, expert) pair obtained by multidimensional scaling (MDS) of the second-order similarity matrix. Each dot represents a single expert from one model run, with colour denoting cluster membership. The boxed points indicate the randomly selected experts used to illustrate each cluster. b) The 10 images eliciting the strongest activation (highest $L_2$ output norm over the full THINGS dataset) for five experts per cluster.} 
  \label{fig:mainfig_stability}
\end{figure}

We observed a bipartitioned representation across both model configurations: one cluster captured animate, organic, and irregularly-textured images, and the other predominantly inanimate, artificial, geometrically-structured images (Figure~\ref{fig:mainfig_stability}). As before, this points to a binary partition of inputs along an animate-inanimate axis---one of the most robust organisational axes in human inferior temporal cortex. The pattern is likely driven by animacy being the largest source of variance in natural image statistics, such that any system optimised for visual discrimination tends to recover it as a principal organisational axis. One might argue that the bipartition stems from training with top-$k = 2$ in the four-expert model, but the same split also predominates in the eight- and sixteen-expert configurations (Figure~\ref{fig:mainfig_stability}), even though top-$k = 2$ does not consistently select the same pair of experts. This suggests that the partition is stable both across model configurations and across independent instantiations.

\section{Discussion}

We implemented a contrastive Mixture-of-Experts Convolutional Neural Network (MoE-CNN) loosely inspired by cortical processing. Trained via self-supervision on image statistics, our MoE-CNN developed tuning profiles spanning a range of features that cannot be fully explained by image category alone. While routing decisions suggest sparse preferences for particular categories, probing of expert tuning revealed a much broader and more diverse set of features to which each expert becomes tuned. Despite this diversity, multivariate analyses reveal a shared representational structure across experts, with animacy emerging as the dominant axis. This distinction is also consistently found across human visual regions \citep{konkle_tripartite_2013, thorat_nature_2019, blumenthal_animacy_2018}. 

Representational similarity analysis further recovered structure that persists across independent model initialisations, pointing to stable visual features underlying stimulus partitioning. Although emphasised feature axes can vary even when experts share similar geometries, some motifs re-emerge reliably, notably the animate–inanimate partition. More broadly, our results suggest that experts are not best characterised as sparsely tuned detectors for discrete categories. Instead, each expert reflects a distributed mixture of continuous, behaviourally relevant dimensions, consistent with contemporary accounts of distributed tuning in human ventral visual cortex. Overall, applying tools and insights from visual neuroscience extends how experts in vision MoEs can be understood, revealing properties of the model that we hope can be leveraged in further interdisciplinary research.

\paragraph{Limitations.} Several aspects of our setup constrain the scope of our findings. First, expert specialisation likely depends on the image statistics of the training set. Our models were trained on a relatively small dataset with predefined categories, yet we still observe specialisations that cut across category boundaries. Training on broader, more naturalistic datasets such as Ecoset \citep{mehrer_ecologically_2021} may yield richer expert representations, opening avenues for using MoEs as data-driven models of cortical organisation in higher-level occipitotemporal cortex. Second, while we focus on self-supervised contrastive learning to keep the learning signal ecologically plausible and free of predefined categorical structure, specialisation patterns may differ across objectives, making a systematic comparison a natural next step. Third, we intentionally keep the model size approximately constant across $E$ to disentangle specialisation from scale, which is atypical in MoE practice, where increasing $E$ usually increases the total number of parameters. Our findings may therefore not extend directly to large-scale industrial MoEs, where greater capacity per expert may influence the representations developed.

\paragraph{Future Avenues.} Our work deepens understanding of the visual properties to which experts in vision MoE models become tuned under a given visual diet, revealing dynamics of expert organisation that extend beyond routing alone. Why experts reliably recover some features, but not others, remains an open question with direct relevance to visual neuroscience. MoE models, therefore, offer a computational framework for addressing long-standing questions about why visual features are distributed across the cortex as they are, and what constraints and pressures give rise to particular patterns of specialisation. At the same time, our study introduces a new perspective for interpreting MoE experts: expert-level probing reveals an expert's internal decomposition of its inputs in ways that routing alone cannot, suggesting that expert specialisation is better characterised as sparse tuning to diverse, continuous features rather than to discrete categorical classes. We hope this perspective offers a useful lens for understanding the internal processing hierarchy of vision MoE networks.

\section{Acknowledgements}

The authors thank Katharina Dobs for the valuable discussion on the research. 

\newpage
\bibliography{moe_neurips}

% \newpage
% \input{checklist.tex}

\newpage

\appendix
\renewcommand{\thefigure}{S\arabic{figure}}
\renewcommand{\thetable}{S\arabic{table}}
\renewcommand{\theHfigure}{S\arabic{figure}}  % fixes hyperref anchors
\renewcommand{\theHtable}{S\arabic{table}}    % fixes hyperref anchors
\setcounter{figure}{0}
\setcounter{table}{0}

\section{Supplementary Methods}

\subsection{Model implementation details}

We trained the MoE-CNN on the unlabeled split of the STL-10 dataset, consisting of 100,000 images. Each input image was transformed into two augmented views following a standard contrastive learning pipeline, including random resized cropping, random horizontal flipping, colour jittering, random grayscale conversion, Gaussian blur, and normalisation using the dataset mean and standard deviation. All images were resized and cropped to a spatial resolution of $96 \times 96$ pixels.

The model consists of a convolutional backbone (ResNet-based) with $5$ blocks, combined with a mixture-of-experts (MoE) module comprising $E \in \{4, 8, 16\}$ experts. A shared prefix of the first three backbone blocks processes each image before expert routing. Each block applies a stride-2 convolution, so the feature map entering the gating network has a spatial resolution of $12 \times 12$. To maintain an approximately equivalent parameter count, channels in the expert-specific blocks are scaled by $1/\sqrt{E}$, giving $C = \lfloor 256/\sqrt{E} \rfloor$ channels at the gating point (128, 90, and 64 for $E = 4, 8, 16$, respectively). A top-$k$ gating network ($k = 2$) maps this $C \times 12 \times 12$ feature map to $E$ routing logits via a $3 \times 3$ convolution, global average pooling, and a linear layer --- compressing the full feature map (e.g., $18{,}432$ values for $E = 4$) into $E$ scalars. The top-2 logits are retained and normalised with softmax to produce sparse expert weights, while all other experts receive zero weight. Each expert produces a $\lfloor 256/\sqrt{E} \rfloor$-dimensional readout representation, which is projected to a fixed 128-dimensional embedding via a 2-layer MLP projection head. Contrastive learning was performed using a temperature-scaled NT-Xent objective with temperature $\tau = 0.5$.

Optimization was performed using the Adam optimizer with an initial learning rate of $3 \times 10^{-4}$ and weight decay of $1 \times 10^{-5}$. The learning rate was scheduled using cosine annealing with a minimum learning rate of $1 \times 10^{-6}$. An auxiliary load-balancing term was included in the loss with weight $0.1$ to encourage balanced expert utilisation.

Models were trained for $100$ epochs with a batch size of $256$. Training was implemented in PyTorch using the PyTorch Lightning framework, with mixed-precision training (16-bit) and gradient clipping (maximum norm of $1.0$). Validation loss was monitored for checkpointing, and the top-$3$ performing models (along with the final model) were retained.

All experiments were conducted on a single NVIDIA V100 GPU (32GB VRAM) with 62 GB system RAM. Training a single model instance (one full training run of 100 epochs on the 2-view augmented STL-10 dataset described above) took approximately 3–4 hours. We trained 3 exemplar model configurations, with 10 additional independent instances per configuration. Total compute was approximately 115.5 GPU-hours.

\subsection{Linear regression analysis of expert responses}

We quantified the extent to which human-interpretable semantic dimensions explain variation in expert routing and response magnitude. For each expert and each image in the THINGS dataset (1,854 object concepts, thus 1,854 representative images), we extracted (i) pre-softmax gating activations (logits) and (ii) the $L_2$ norm of the readout activations. 

Each image was associated with a 66-dimensional embedding derived from the THINGS dataset, where each dimension reflects a behaviorally derived semantic attribute. These embeddings served as predictors $X \in \mathbb{R}^{1854 \times 66}$, while the model responses (gating logits or readout norms) formed the targets $y \in \mathbb{R}^{1854}$.

We fit a non-negative L1-regularised linear model (Lasso) using the \texttt{scikit-learn} implementation. To obtain unbiased estimates of predictive performance while tuning regularisation, we employed nested cross-validation. The inner loop (5-fold) performed hyperparameter selection via grid search over 40 regularisation strengths sampled logarithmically in the range $[10^{-4}, 10^{0}]$.

The optimal regularisation parameter selected in the inner loop was then used to fit the model within each training split of the outer loop. The outer loop (5-fold) provided an unbiased estimate of generalisation performance, reported as mean $\pm$ standard deviation of the coefficient of determination ($r^2$) across folds (Table 1).

Finally, model coefficients were obtained by refitting the Lasso model on the full dataset using the optimal regularisation parameter. For interpretability, we report the top semantic dimensions with the largest regression coefficients for each expert.

\subsection{Logistic regression for category separability}

We quantified each expert's class separability using logistic regression (\texttt{scikit-learn}). For the STL-10 test set, we extracted readout activations from each expert across all labelled images and trained a binary logistic classifier on every pair of the 10 categories. The regularisation parameter was kept at its default ($C = 1.0$), and we report 5-fold cross-validated balanced accuracy, where 0.5 corresponds to chance and 1.0 to perfect linear separability.

We generalised this analysis to the THINGS dataset, which provides $\sim$26k images spanning 1,854 concepts that can be grouped into 27 higher-level core categories \citep{hebart_things_2019}. Since some concepts map to multiple or no core categories, we restricted the analysis to concepts with a single category label, pooling all images within a concept to increase sample size. We then ran the same 5-fold cross-validated pairwise logistic regression across the 27 categories, again reporting balanced accuracy.

\subsection{Representational similarity analysis}

For our first-order representational similarity analysis (RSA), we chose Euclidean over cosine distance because Euclidean distance preserves response-magnitude differences that cosine would discard. RSA is well-suited to comparing representations across independently trained networks, as it is invariant to rotation and isotropic scaling. We used the THINGS Similarity dataset rather than STL10 for two reasons: its 1,854 reference images keep the first-order RDM computation tractable, while its broad coverage of an everyday visual diet provides richer and more semantically diverse stimuli than ten coarse categories.

\section{Supplementary Results}

\subsection{Expert routing consistency}

During training, we imposed no explicit constraint requiring augmented views of the same image to be routed to the same pair of experts under $k = 2$. When paired views are routed to different expert sets, the experts are pushed toward more generalist, overlapping representations. We found that enforcing such a constraint explicitly made the model substantially harder to converge. We instead examined whether routing consistency emerges naturally as a by-product. To quantify this, we computed the \emph{top-2 agreement}: the fraction of images for which both augmented views activate the same pair of experts (across $n = 10$ per model configuration). The mean top-2 agreement was 79.27\% ($\pm$ 2\%), 58.32\% ($\pm$ 4\%), 45.30\% ($\pm$ 2\%), respectively, all well above the corresponding chance levels of $1/\binom{E}{2}$, namely 16.66\%, 3.57\%, and 0.83\%. This indicates that the model spontaneously learns to route augmented images from the same origin to a reasonably consistent set of experts without the need for explicit pressure to do so.

\section{Supplementary Figures and Tables}

\begin{figure}[!htpb]
  \centering
  \includegraphics[width=0.75\linewidth]{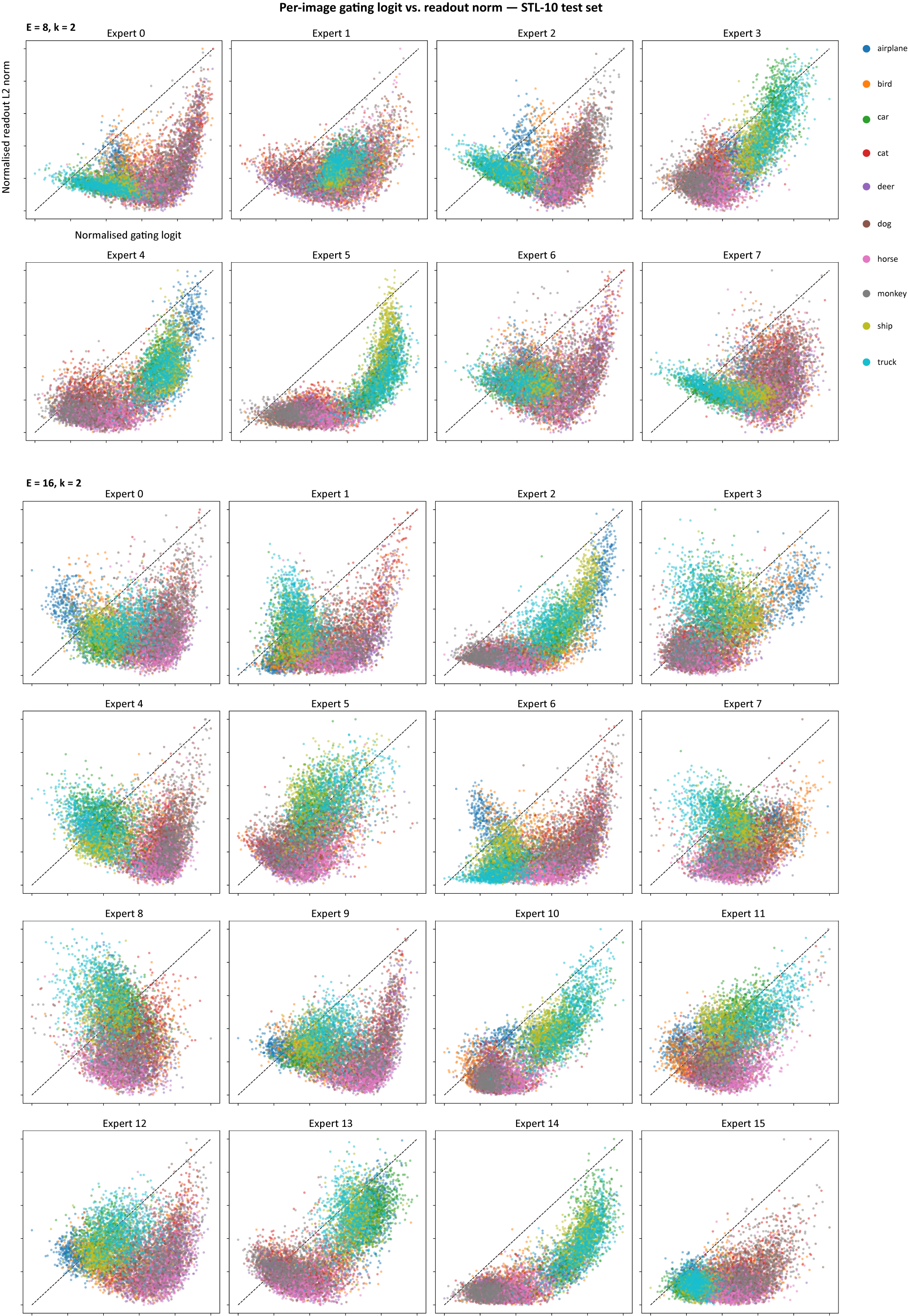}
  \caption{Scatterplot showing the relationship between gating logits and readout activation norms across the held-out STL10 test set (8k samples; 800 images per class). Min--max-normalised gating logits are plotted against normalised readout response magnitudes for each expert across models with $E \in \{8, 16\}$ and $k = 2$. Points are coloured by image class. The dashed diagonal indicates the relationship that would be expected if increases in gating logits always corresponded to increases in readout magnitude.}
  \label{fig:supplementary_gating_vs_readout}
\end{figure}

\begin{figure}[!htpb]
  \centering
  \includegraphics[width=\linewidth]{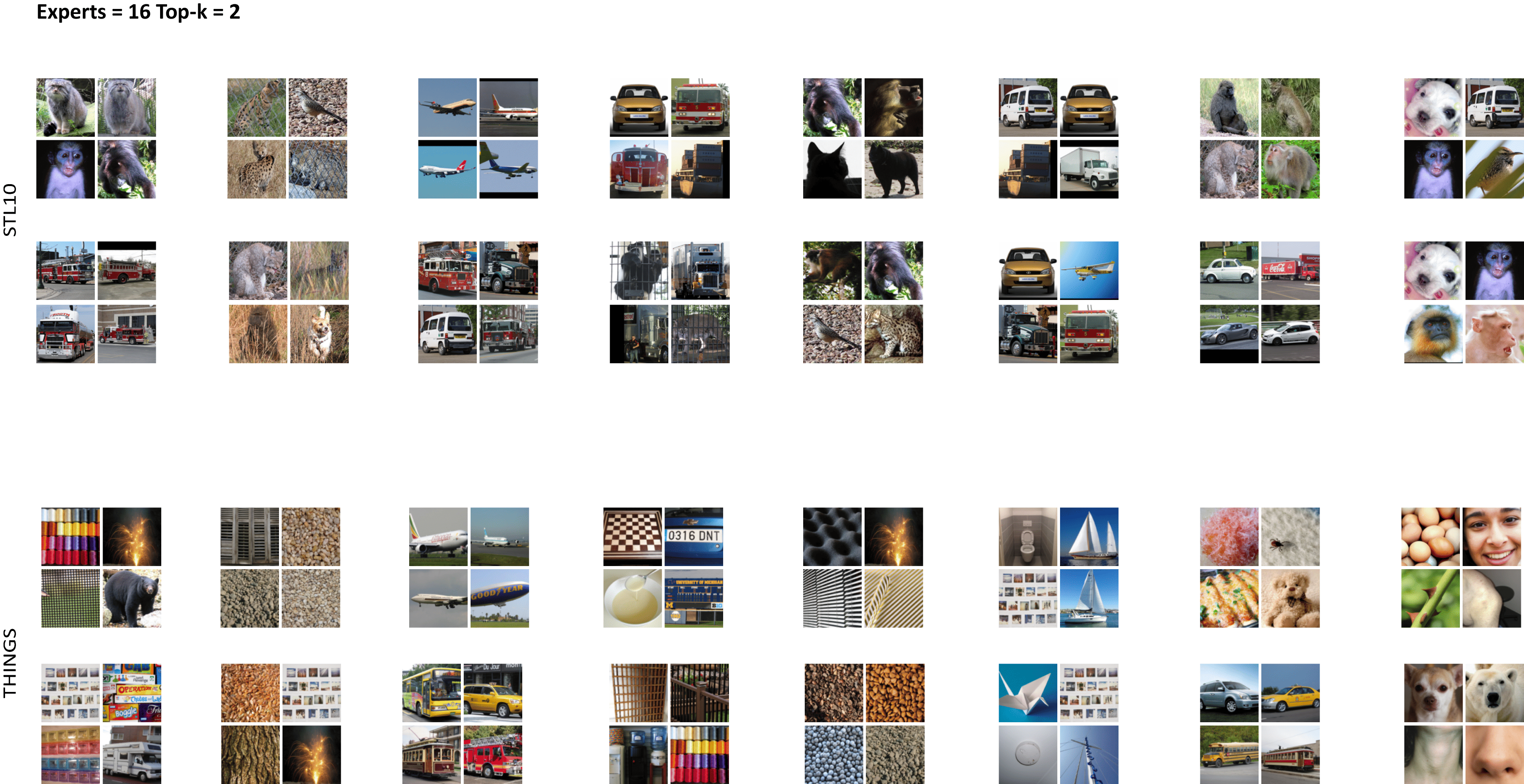}
  \caption{Images eliciting the highest response magnitudes at the readout layer for each expert in a sixteen-expert model. The top four images for each expert from the held-out STL10 test set (top) and THINGS dataset (bottom) are shown.}
  \label{fig:supplementary_mei}
\end{figure}

\begin{table}[!htpb]
\caption{Loadings of behaviour-derived dimensions on each expert's gating profile (which inputs route to the expert) and readout response-magnitude profile (what the expert encodes), for the eight- and sixteen-expert models. Each row reports the expert index, the $r^2$ from a five-fold cross-validated non-negative L1 regression, and the top five dimensions with their corresponding coefficients.}
\centering
\label{tab:supplementary_things_loadings}
\scriptsize
\setlength{\tabcolsep}{3pt}
\renewcommand{\arraystretch}{1.1}
\resizebox{\linewidth}{!}{%
\begin{tabular}{@{}c c c p{2.6cm} c p{2.6cm} c p{2.6cm} c p{2.6cm} c p{2.6cm} c@{}}
\toprule
Expert & $r^2$ Mean & $r^2$ Std & Feature 1 & Weight & Feature 2 & Weight & Feature 3 & Weight & Feature 4 & Weight & Feature 5 & Weight \\
\midrule
\multicolumn{13}{l}{\textit{$E = 8$, k = 2 — Gating}} \\
\midrule
0 & .526 & .016 & animal-related & 0.335 & repetitive/spiky & 0.267 & powdery/earth-related & 0.259 & food-related & 0.231 & outdoors & 0.143 \\
1 & .396 & .033 & body-/people-related & 0.268 & metallic/artificial & 0.253 & flying-/sky-related & 0.213 & stick-shaped/container & 0.169 & yellow & 0.165 \\
2 & .374 & .018 & animal-related & 0.262 & food-related & 0.169 & plant-related & 0.143 & transparent/shiny/crystalline & 0.120 & body-/people-related & 0.105 \\
3 & .511 & .024 & house-related/furnishing-related & 0.222 & transportation-/movement-related & 0.207 & electronics/technology & 0.151 & oriented/many things & 0.112 & box-related/container & 0.111 \\
4 & .545 & .029 & flying-/sky-related & 0.303 & house-related/furnishing-related & 0.300 & metallic/artificial & 0.252 & long/thin & 0.208 & electronics/technology & 0.186 \\
5 & .449 & .014 & house-related/furnishing-related & 0.269 & transportation-/movement-related & 0.229 & electronics/technology & 0.154 & paper-related/flat & 0.121 & box-related/container & 0.120 \\
6 & .489 & .012 & animal-related & 0.346 & repetitive/spiky & 0.312 & powdery/earth-related & 0.293 & food-related & 0.242 & outdoors & 0.193 \\
7 & .449 & .039 & animal-related & 0.260 & food-related & 0.216 & powdery/earth-related & 0.205 & body part-related & 0.169 & long/thin & 0.105 \\
\midrule
\multicolumn{13}{l}{\textit{$E = 8$, k = 2 — Readout}} \\
\midrule
0 & .111 & .060 & fine-grained pattern & 0.258 & repetitive/spiky & 0.229 & animal-related & 0.174 & body part-related & 0.108 & tubular & 0.095 \\
1 & .137 & .048 & black & 0.157 & electronics/technology & 0.137 & body-/people-related & 0.121 & stick-shaped/container & 0.119 & oriented/many things & 0.116 \\
2 & .088 & .033 & body part-related & 0.193 & flying-/sky-related & 0.173 & black & 0.134 & repetitive/spiky & 0.122 & tubular & 0.107 \\
3 & .283 & .044 & electronics/technology & 0.221 & transportation-/movement-related & 0.199 & house-related/furnishing-related & 0.139 & stick-shaped/container & 0.123 & upright/elongated/volumous & 0.118 \\
4 & .239 & .019 & flying-/sky-related & 0.188 & electronics/technology & 0.177 & house-related/furnishing-related & 0.146 & stick-shaped/container & 0.129 & black & 0.110 \\
5 & .196 & .032 & house-related/furnishing-related & 0.158 & transportation-/movement-related & 0.142 & electronics/technology & 0.142 & fine-grained pattern & 0.087 & body part-related & 0.072 \\
6 & .131 & .048 & fine-grained pattern & 0.386 & repetitive/spiky & 0.250 & grid-/grating-related & 0.160 & black & 0.106 & animal-related & 0.104 \\
7 & .085 & .032 & body part-related & 0.179 & animal-related & 0.161 & black & 0.151 & repetitive/spiky & 0.137 & tubular & 0.135 \\
\midrule
\multicolumn{13}{l}{\textit{$E = 16$, k = 2 — Gating}} \\
\midrule
0 & .253 & .032 & animal-related & 0.346 & upright/elongated/volumous & 0.256 & powdery/earth-related & 0.164 & food-related & 0.163 & repetitive/spiky & 0.161 \\
1 & .472 & .014 & repetitive/spiky & 0.462 & animal-related & 0.375 & powdery/earth-related & 0.303 & food-related & 0.242 & outdoors & 0.194 \\
2 & .443 & .027 & flying-/sky-related & 0.419 & house-related/furnishing-related & 0.320 & paper-related/flat & 0.262 & tools/handheld/elongated & 0.208 & tubular & 0.199 \\
3 & .346 & .024 & flying-/sky-related & 0.406 & tools/handheld/elongated & 0.208 & paper-related/flat & 0.170 & tubular & 0.163 & body part-related & 0.159 \\
4 & .396 & .014 & animal-related & 0.327 & food-related & 0.212 & powdery/earth-related & 0.159 & transparent/shiny/crystalline & 0.154 & black & 0.129 \\
5 & .509 & .026 & upright/elongated/volumous & 0.259 & house-related/furnishing-related & 0.227 & long/thin & 0.182 & body-/people-related & 0.179 & fluid-related/drink-related & 0.174 \\
6 & .553 & .019 & animal-related & 0.394 & powdery/earth-related & 0.285 & food-related & 0.276 & repetitive/spiky & 0.240 & coarse pattern/many things & 0.120 \\
7 & .307 & .026 & flying-/sky-related & 0.263 & body part-related & 0.227 & tools/handheld/elongated & 0.207 & food-related & 0.204 & spherical/voluminous & 0.193 \\
8 & .290 & .022 & food-related & 0.223 & tools/handheld/elongated & 0.186 & animal-related & 0.177 & powdery/earth-related & 0.160 & paper-related/flat & 0.149 \\
9 & .361 & .033 & animal-related & 0.445 & repetitive/spiky & 0.399 & powdery/earth-related & 0.282 & food-related & 0.261 & valuable/precious & 0.241 \\
10 & .402 & .036 & house-related/furnishing-related & 0.258 & transportation-/movement-related & 0.245 & oriented/many things & 0.178 & electronics/technology & 0.175 & box-related/container & 0.157 \\
11 & .300 & .039 & transportation-/movement-related & 0.181 & upright/elongated/volumous & 0.174 & oriented/many things & 0.170 & house-related/furnishing-related & 0.160 & valuable/precious & 0.158 \\
12 & .356 & .036 & animal-related & 0.422 & repetitive/spiky & 0.398 & powdery/earth-related & 0.258 & food-related & 0.258 & valuable/precious & 0.252 \\
13 & .408 & .022 & flying-/sky-related & 0.201 & house-related/furnishing-related & 0.190 & transportation-/movement-related & 0.184 & paper-related/flat & 0.183 & tubular & 0.179 \\
14 & .472 & .016 & house-related/furnishing-related & 0.325 & transportation-/movement-related & 0.270 & electronics/technology & 0.194 & metallic/artificial & 0.137 & box-related/container & 0.129 \\
15 & .325 & .050 & animal-related & 0.254 & food-related & 0.157 & body-/people-related & 0.138 & spherical/voluminous & 0.135 & upright/elongated/volumous & 0.130 \\
\midrule
\multicolumn{13}{l}{\textit{$E = 16$, k = 2 — Readout}} \\
\midrule
0 & .170 & .033 & house-related/furnishing-related & 0.199 & upright/elongated/volumous & 0.198 & stick-shaped/container & 0.172 & body part-related & 0.160 & black & 0.148 \\
1 & .220 & .065 & repetitive/spiky & 0.300 & fine-grained pattern & 0.298 & grid-/grating-related & 0.188 & animal-related & 0.170 & transportation-/movement-related & 0.129 \\
2 & .256 & .071 & house-related/furnishing-related & 0.255 & electronics/technology & 0.229 & transportation-/movement-related & 0.143 & stick-shaped/container & 0.142 & medicine-/health-related & 0.117 \\
3 & .275 & .042 & house-related/furnishing-related & 0.229 & electronics/technology & 0.216 & flying-/sky-related & 0.178 & body part-related & 0.171 & black & 0.161 \\
4 & .161 & .026 & body part-related & 0.208 & black & 0.197 & electronics/technology & 0.137 & stick-shaped/container & 0.130 & yellow & 0.112 \\
5 & .297 & .042 & house-related/furnishing-related & 0.242 & electronics/technology & 0.241 & stick-shaped/container & 0.199 & yellow & 0.164 & black & 0.135 \\
6 & .250 & .028 & body part-related & 0.280 & fine-grained pattern & 0.273 & animal-related & 0.227 & food-related & 0.163 & tubular & 0.157 \\
7 & .220 & .048 & body part-related & 0.310 & house-related/furnishing-related & 0.232 & electronics/technology & 0.201 & white & 0.183 & yellow & 0.168 \\
8 & .103 & .036 & fine-grained pattern & 0.244 & body part-related & 0.209 & repetitive/spiky & 0.181 & transportation-/movement-related & 0.170 & electronics/technology & 0.152 \\
9 & .097 & .043 & fine-grained pattern & 0.241 & body part-related & 0.159 & repetitive/spiky & 0.152 & grid-/grating-related & 0.115 & animal-related & 0.109 \\
10 & .333 & .020 & electronics/technology & 0.228 & house-related/furnishing-related & 0.219 & transportation-/movement-related & 0.218 & stick-shaped/container & 0.113 & yellow & 0.103 \\
11 & .286 & .028 & electronics/technology & 0.234 & house-related/furnishing-related & 0.211 & upright/elongated/volumous & 0.175 & oriented/many things & 0.174 & transportation-/movement-related & 0.165 \\
12 & .119 & .046 & repetitive/spiky & 0.294 & fine-grained pattern & 0.181 & black & 0.140 & transportation-/movement-related & 0.128 & electronics/technology & 0.104 \\
13 & .282 & .024 & electronics/technology & 0.247 & house-related/furnishing-related & 0.230 & transportation-/movement-related & 0.172 & flying-/sky-related & 0.155 & yellow & 0.133 \\
14 & .350 & .025 & house-related/furnishing-related & 0.231 & electronics/technology & 0.217 & transportation-/movement-related & 0.176 & stick-shaped/container & 0.156 & long/thin & 0.150 \\
15 & .163 & .013 & body part-related & 0.145 & white & 0.127 & stick-shaped/container & 0.102 & animal-related & 0.091 & body-/people-related & 0.086 \\
\bottomrule
\end{tabular}%
}
\end{table}

\begin{figure}[!htpb]
  \centering
  \includegraphics[width=\linewidth]{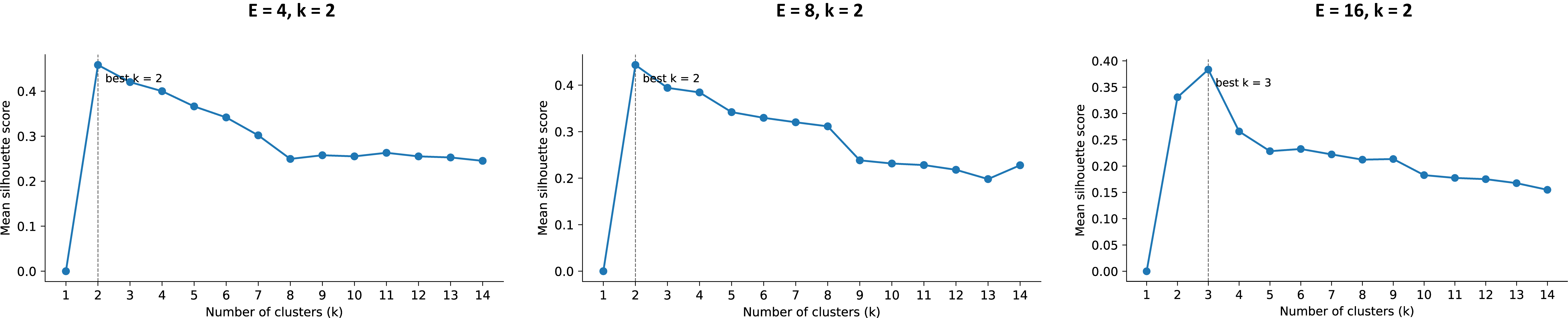}
  \caption{Mean silhouette score as a function of the number of clusters used to select the optimal clustering for four- (left) and eight- (middle), and sixteen-expert models (right).}
  \label{fig:supplementary_scree}
\end{figure}

\begin{figure}[!htpb]
  \centering
  \includegraphics[width=\linewidth]{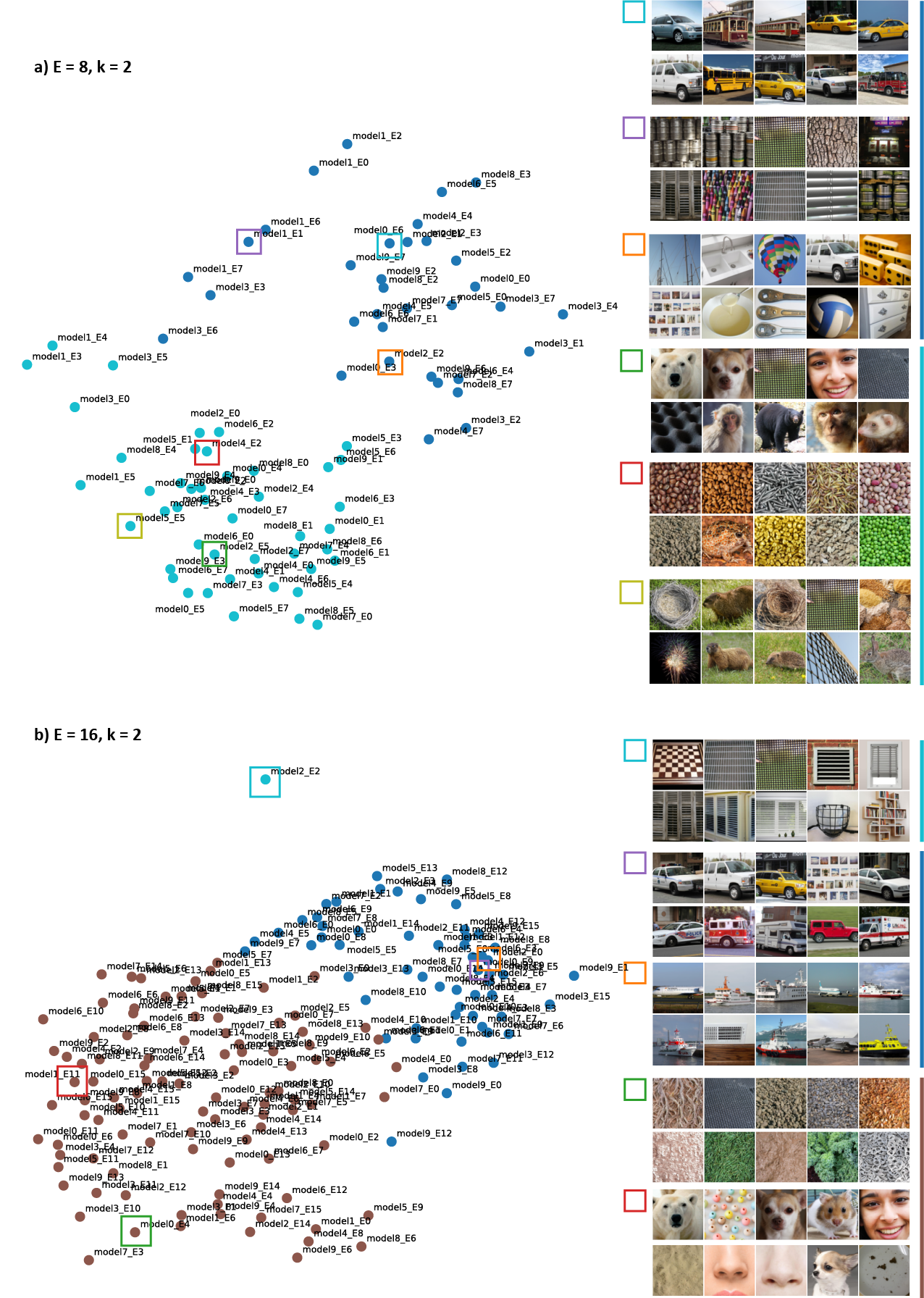}
    \caption{Consistency of expert specialisations across 10 independently trained model instantiations for (a) eight- and (b) sixteen-expert models. Left: Two-dimensional MDS projection of the second-order similarity matrix over (model, expert) pairs. Each dot represents one expert from one model run, coloured by cluster membership. Boxed points mark the experts randomly selected to illustrate each cluster on the right. Right: The 10 THINGS images eliciting the strongest activation (highest $L_2$ output norm over the full THINGS dataset) for each selected expert.}
    \label{fig:supplementary_stability}
\end{figure}

\newpage

\end{document}